%% file: conference_041818.tex
\documentclass[conference]{IEEEtran}
\IEEEoverridecommandlockouts
\usepackage{cite}
\usepackage{amsmath,amssymb,amsfonts}
\usepackage{algorithmic}
\usepackage{graphicx}
\usepackage{listings}

\usepackage{textcomp}
\usepackage{xcolor}
\def\BibTeX{{\rm B\kern-.05em{\sc i\kern-.025em b}\kern-.08em
    T\kern-.1667em\lower.7ex\hbox{E}\kern-.125emX}}
\begin{document}

\title{Word Embedding Neural Networks to Advance Knee Osteoarthritis Research}

\makeatletter
\newcommand{\linebreakand}{%
  \end{@IEEEauthorhalign}
  \hfill\mbox{}\par
  \mbox{}\hfill\begin{@IEEEauthorhalign}
}
\makeatother

\author{\IEEEauthorblockN{Soheyla Amirian} 
\IEEEauthorblockA{\text{University of Georgia}\\amirian@uga.edu}
\and\IEEEauthorblockN{Husam Ghazaleh} 
\IEEEauthorblockA{
\text{Quincy University}\\ghazahu@quincy.edu}
\and\IEEEauthorblockN{Mehdi Assefi} 
\IEEEauthorblockA{\text{University of Georgia}\\asf@uga.edu}
\and\IEEEauthorblockN{Hilal Maradit Kremers} 
\IEEEauthorblockA{\text{Mayo Clinic}\\maradit@mayo.edu}
\linebreakand
\IEEEauthorblockN{Hamid R. Arabnia} 
\IEEEauthorblockA{\text{University of Georgia}\\hra@uga.edu}
\and\IEEEauthorblockN{Johannes F. Plate} 
\IEEEauthorblockA{\text{University of Pittsburgh}\\johannes.plate@pitt.edu}
\and\IEEEauthorblockN{Ahmad P. Tafti} 
\IEEEauthorblockA{\text{University of Pittsburgh}\\tafti.ahmad@pitt.edu}
}

\maketitle

\begin{abstract}
\input{tex/0_Abstract}
\end{abstract}


\section{Introduction}
\label{sec:lit}
\input{tex/1_inro}

\section{Materials and Methods}
\label{sec:MM}
\input{tex/2_MM}

\section{Experimental Validation and Scientific Visualization}
\label{sec:EV}
\input{tex/3_EV}

\section{Conclusion and Outlook}
\label{sec:Conclution}
\input{tex/4_Conclusion.tex}

\section*{Acknowledgment}

The authors declare that they have no competing interests.

\end{document}

%% file: tex/0_Abstract.tex
Osteoarthritis (OA) is the most prevalent chronic joint disease worldwide, where knee OA takes more than 80\% of commonly affected joints. Knee OA is not a curable disease yet, and it affects large columns of patients, making it costly to patients and healthcare systems. Etiology, diagnosis, and treatment of knee OA might be argued by variability in its clinical and physical manifestations. Although knee OA carries a list of well-known terminology aiming to standardize the nomenclature of the diagnosis, prognosis, treatment, and clinical outcomes of the chronic joint disease, in practice there is a wide range of terminology associated with knee OA across different data sources, including but not limited to biomedical literature, clinical notes, healthcare literacy, and health-related social media. Among these data sources, the scientific articles published in the biomedical literature usually make a principled pipeline to study disease. Rapid yet, accurate text mining on large-scale scientific literature may discover novel knowledge and terminology to better understand knee OA and to improve the quality of knee OA diagnosis, prevention, and treatment. The present works aim to utilize artificial neural network strategies to automatically extract vocabularies associated with knee OA diseases. Our finding indicates the feasibility of developing word embedding neural networks for autonomous keyword extraction and abstraction of knee OA.

%% file: tex/1_inro.tex
Osteoarthritis (OA) is the most prevalent chronic joint disease worldwide, where knee OA takes more than 80\% of commonly affected joints \cite{rf1, rf2}. A vast amount of biomedical research articles are published daily, accumulating rich information about the disease, including but not limited to knee OA diagnosis, prognosis, and treatment. Rapid yet, little is done to automatically analyze this large column of scientific articles to first better understand the disease and then to improve the quality of patient care, enabling healthcare providers to perform clinical practice with autonomous and real-time updates from new articles published in the literature.

Of late, healthcare systems are widely applying advanced technologies, such as artificial intelligence (AI) and machine learning to boost medical processes \cite{rf3, rf4, rf5, rf6}. Recently, we have had widespread applications of AI-supported technologies in healthcare institutions to improve care service quality and efficiency of medical resources. AI-supported technologies learn and diagnose from a large volume of medical research and patients’ treatment records. 
Therefore, AI has attracted the attention of researchers, physicians, technology and program developers, and consumers in various fields because of its potential for transformative innovations in treating human diseases and public health. Among all AI strategies, natural language processing (NLP) combined with machine learning algorithms and word embeddings have demonstrated successful applications in harnessing unstructured text data, and particularly word embeddings have been added to clinical text analytics as a powerful representation model to address a variety of problems in named entity recognition \cite{rf7,rf8}, text classification \cite{rf9}, and text summarization \cite{rf10, rf11}. Inspired by the current advances in applying word embeddings in different healthcare informatics problems, the present work aims to train and validate a widely-used word embedding neural network, namely Word2Vec \cite{rf12, rf13} on top of the large body of scientific articles available at PubMed \cite{rf14} to automatically detect, identify, and characterize terms relevant to knee OA. The main significance of our work is listed as follows:

%\textcolor{blue}{NN and Word embedding}\\

\begin{itemize}
\setlength\itemsep{0.5em}
  \item \textbf{Clinical Significance:} First, the current work demonstrates the potential for building, training, and validating word embedding neural networks using scientific articles published in knee OA settings, with a real-time update from new articles published on a daily basis. Second, we built a computational framework to automatically collect PubMed data and assemble a retrospective textual dataset that can advance open scientific research in epidemiology, etiology, diagnosis, and treatment of knee OA.
  
  \item \textbf{Technical Significance:} First, we proposed and implemented an automated word embedding neural network pipeline that fits well in extracting vocabularies and keywords associated with knee OA disease, utilizing scientific articles. Second, we proposed a word embedding neural network pipeline that effectively identifies keyword proximity and disease terminology in a clinical setting. 
 
 \end{itemize}
The organization of the paper is as follows. Section II discusses the materials and methods. Experimental validation and scientific visualization are presented in Section III. Section IV concludes the study and draws future directions.

 \begin{figure*}[!ht]
    \centering
    \includegraphics[width=\linewidth]{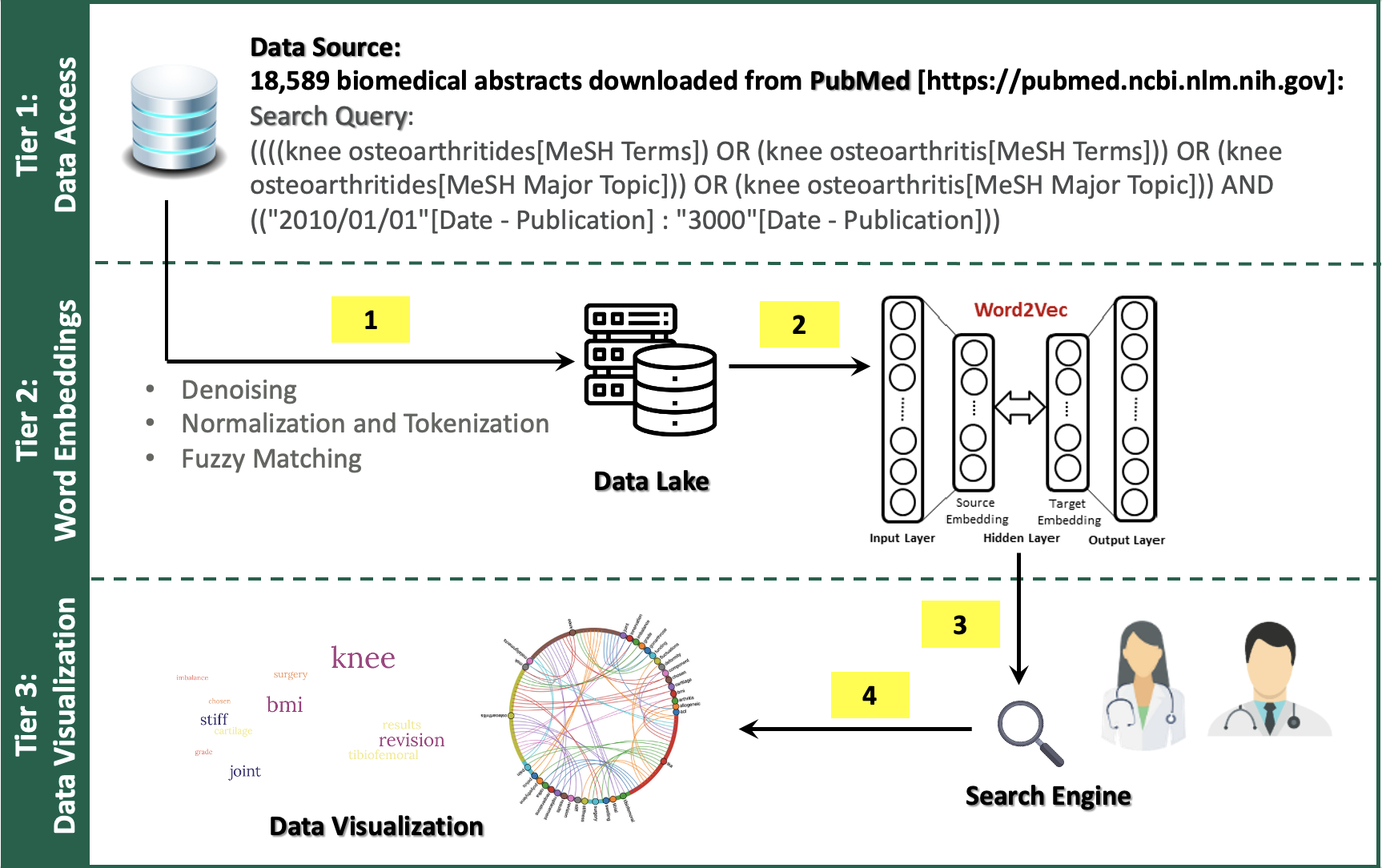}
    \caption{The proposed software architectural model to extract keyword similarities in knee OA context using word2vec word embedding.} 
    \label{2_figv2}
    \end{figure*}

%% file: tex/2_MM.tex
To make this work self-contained, we start with word embeddings. Word embeddings refer to those computational text mining methods that map each word in a given text data to a vector of real numbers. For example, applying word2vec word embedding on a given medical text corpus (e.g., clinical notes), represents the word ``tibiofemoral'' as a vector of [0.42, 0.61, -1.37, 1.29, 0.63]. This representation fits well with artificial neural nets since their architectures only process continuous numbers and not characters. There are two methods in word embeddings: (1) {\it context-based},  which basically falls into supervised learning algorithms, where given text data, the method builds a predictive model to predict the target words, and (2) {\it count-based}, which mainly accounts for word frequency and it works as an unsupervised learning algorithm \cite{rf15}.

\subsection{Word2vec}
Word2vec \cite{rf12, rf13} word embeddings learns geometrical vectors of words within a document. Instead of only capturing the word intensities across a textual dataset, they also capture word order, higher-level syntax, and semantics. Generally speaking, the main purposes of word embeddings are: (1) to create an input for machine learning algorithms (e.g., as a set of features), (2) to find nearest neighbors in the embedding space, and (3) to do data visualization and relation extraction among words. Word2vec is classified into two learning mechanisms, namely continuous bag-of-words (CBOW) and skip-gram models. The CBOW predicts a target word given a context, while conversely, skip-gram will predict a target context given a word. They then try to minimize a well-defined loss function (e.g., hierarchical softmax, full softmax, or noise contrastive estimation). For instance, utilizing the skip-gram model, one loss function could be the full softmax, thus the very final output layer will apply softmax to estimate the probability of predicting the output word $W_{out}$ given $W_{in}$, as follows:

\begin{equation}
 P(W_{out}|W_{in}) = \frac{\exp ({{v'_{W_{out}}}}^T {{v_{W_{in}}}})}{\sum_{i=1}^{V} \exp ({{v'_{W_{i}}}}^T v_{W_{in}}) }
\end{equation}

Here, the embedding vector of every single word is defined by the matrix $W$, and the context vector is specified by the output matrix $W'$. Given an input word as $W_{in}$, we label the corresponding row of the matrix $W$ as vector $v_{W_{in}}$, the embedding vector, and its corresponding column of $W'$ as $v'_{W_{in}}$, the context vector.

\subsection{Proposed Software Architectural Model}

Figure~\ref{2_figv2} demonstrates the proposed processing pipeline to extract keyword similarities relevant to knee OA disease automatically. This architectural model spans three different tiers: (1) Data access, (2) Word embeddings, and (3) Data visualization. 

Tier 1 downloads the biomedical abstracts from PubMed, and it does text pre-processing steps, such as document denoising (e.g., deleting email addresses, digits, and characters including ``['', ``]'', "\%"), document tokenization and normalization and finally stop word elimination and fuzzy matching. Tier 1 then sends the words to Tier 2 to train the word embeddings. Once the model has been trained, then given a term, such as ``osteoarthritis'' in the search engine, the average of the highest cosine distance value in the learning models generated by word2vec will be measured, and the rest will be data and scientific visualizations. 
 
\subsection{Data Collection and Dataset}
We computationally assembled a dataset using PubMed advanced search \cite {rf16}. We downloaded 18,589 scientific abstracts published within the knee OA setting. The pre-defined MeSH Terms and MeSH Major Topic, including knee osteoarthritides, and knee osteoarthritis have been used to collect relevant abstracts. The query used to download the scientific abstracts was as below: 

\begin{lstlisting}[basicstyle=\ttfamily]
((((knee osteoarthritides[MeSH Terms]) 
OR (knee osteoarthritis[MeSH Terms])) 
OR (knee osteoarthritides[MeSH Major Topic])) 
OR (knee osteoarthritis[MeSH Major Topic])) 
AND (("2010/01/01"[Date - Publication] : 
"3000"[Date - Publication]))
\end{lstlisting}

%% file: tex/3_EV.tex
\begin{figure*}
\centering
    \includegraphics[height=10cm, width=10cm]
    {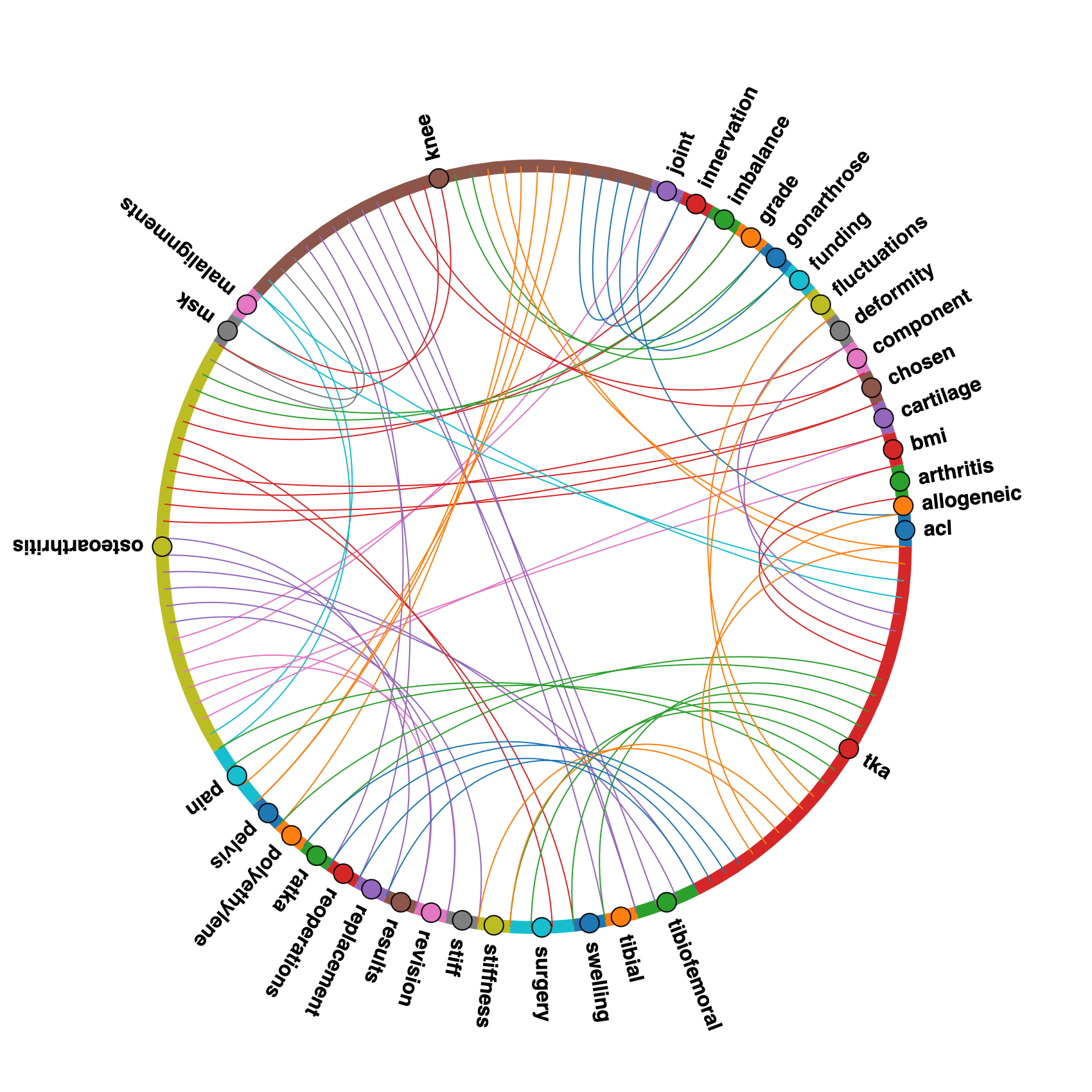}
    \caption{The scientific visualization results were obtained by searching three different terms, including ``osteoarthritis'', ``knee'', and ``tka''. One can see, searching the term ``osteoarthritis'' resulted in clinical meaningful and osteoarthritis-related keywords, such as ``knee'', ``bmi'', ``joint'', ``stiff'', and ``tibiofemoral'' for example. It also resulted in a term such as ``results'', where we could consider it as a noise and/or limitation of the proposed method.} 
    \label{fig:visfig1}
\end{figure*}
\begin{figure*}
\centering
\includegraphics[height=9cm, width=9cm]{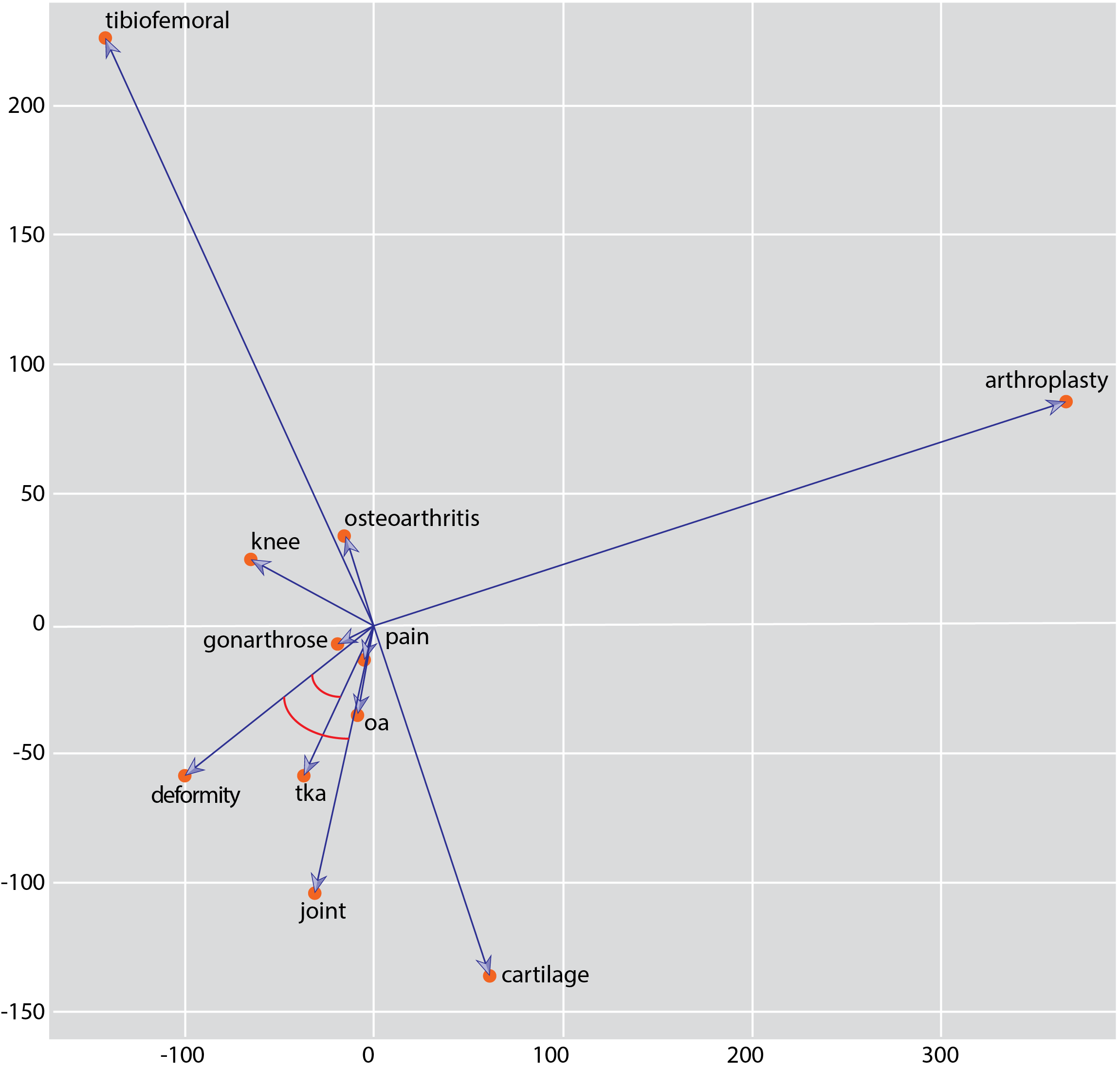}
\caption{The scientific visualization results obtained by searching a list of related terms, as  words = ['knee', 'arthroplasty', 'osteoarthritis', 'tka', 'joint', 'tibiofemoral', 'osteoarthritis', 'deformity', 'gonarthrose', 'cartilage', 'oa', 'pain']. This figure demonstrates the measure of similarity between a set of non-zero vectors of words, and it demonstrates the cosine of the angle between those non-zero vectors. For example, one can see that ``pain'' and ``or'' has very tiny angle between their vectors.} 
\label{fig:visfig2}
\end{figure*}

From the computational perspective, Google Colaboratory or Google Colab equipped with Python 3.8.15 was used to build and run the models. From the experimental validation of the terminology, we used the disease keywords and terminologies available at the Centers for Disease Control and Prevention (CDC) \cite{rf17} as a reference. With that, an average cosine similarity bigger than 0.44 tends to provide better inter-rater reliability among the proposed computational method and the current disease keywords and terminologies available at the Centers for Disease Control and Prevention (CDC) \cite{rf17}. A set of scientific visualizations along with the achieved inter-rater reliabilities between the domain experts and the proposed method are shown in Figure~\ref{fig:visfig1} and Figure~\ref{fig:visfig2}.

\vspace{.1cm}
\vspace{.1cm}

\vspace{.1cm}
\vspace{.1cm}

%% file: tex/4_Conclusion.tex
The scientific abstracts and full articles published in PubMed truly contain recent findings and insights in both clinical and research perspectives. However, the automatic extraction of relevant, clinical meaningful, qualitative terms and keywords from its free-text data is difficult due to complex structure and style of such text data. Keyword extraction for clinical settings is needed to for example summarize the informative text and improve any disease dictionary. This study proposed the integration of a word embedding neural network with knee OA-related abstracts downloaded from PubMed to automatically extract terms and keywords associated with knee OA. The results show the feasibility of using the word embedding neural network in this health-related application area. The results also demonstrated a practical application of our proposed model in extracting important keywords from PubMed abstracts.

Even though the proposed computational method is relatively simple to design and implement, there exists a list of limitations to this study, including but not limited to 1) if the word embedding model did not see a keyword or term in the given corpus, thus it would not be able to generate and then understand a vector representation for such a term, and 2) a cross-lingual utilization of the method is not feasible. Our future works will focus on applying the current method to other musculoskeletal diseases and disorders. We will also incorporate additional word embedding algorithms (e.g., Glove, BERT) into the proposed system and investigate how ensemble word embeddings will tackle the problem. 